\documentclass[11pt,a4paper]{article}
\usepackage[hyperref]{acl2021}
\usepackage{amssymb,graphicx}

\usepackage{times}
\usepackage{latexsym}

\usepackage{microtype}
\usepackage{amsfonts}
\usepackage[utf8]{inputenc}

\usepackage{booktabs}
\usepackage{multirow}
\usepackage{makecell}
\usepackage{bm}
\usepackage{stackengine}
\usepackage{amsmath}

\aclfinalcopy %
\def\aclpaperid{2838} %

\title{KM-BART: Knowledge Enhanced Multimodal BART for Visual Commonsense Generation}

\author{Yiran Xing\thanks{~~The first three authors contribute equally to this work.}$^{~~\spadesuit}$ \quad Zai Shi$^{*\clubsuit}$ \quad Zhao Meng$^{*\clubsuit}$ \\ \quad \textbf{Gerhard Lakemeyer}$^{\spadesuit}$ \quad \textbf{Yunpu Ma}$^{\blacklozenge}$ \quad \textbf{Roger Wattenhofer}$^{\clubsuit}$ \\
  $^\spadesuit$RWTH Aachen, Germany \\
  $^\blacklozenge$LMU Munich, Germany \\
  $^\clubsuit$ETH Zurich, Switzerland \\
  \texttt{\{yiran.xing, gerhard\}@rwth-aachen.de} \\
    \texttt{cognitive.yunpu@gmail.com} \\
    \texttt{\{zaishi, zhmeng, wattenhofer\}@ethz.ch} 
  \\}

\begin{document}
\maketitle
\begin{abstract}

We present \textbf{K}nowledge Enhanced \textbf{M}ultimodal \textbf{BART} (KM-BART), which is a Transformer-based sequence-to-sequence model capable of reasoning about commonsense knowledge from multimodal inputs of images and texts. We adapt the generative BART architecture~\cite{bart} to a multimodal model with visual and textual inputs. We further develop novel pretraining tasks to improve the model performance on the Visual Commonsense Generation (VCG) task. In particular, our pretraining task of Knowledge-based Commonsense Generation (KCG) boosts model performance on the VCG task by leveraging commonsense knowledge from a large language model pretrained on external commonsense knowledge graphs. To the best of our knowledge, we are the first to propose a dedicated task for improving model performance on the VCG task. Experimental results show that our model reaches state-of-the-art performance on the VCG task~\cite{vcg} by applying these novel pretraining tasks.

\end{abstract}

\section{Introduction}

Early work on Vision-Language models has been largely focused on pure understanding tasks~\cite{lxmert, vilbert}.
These models, although improving model performance on understanding tasks such as Visual Question Answering~\cite{vqa}, are not capable of multimodal generation tasks~\cite{caption}.
To ease this problem, researchers have proposed various models~\cite{vlp, oscar} for generating texts based on visual inputs.

These models are mainly pretrained on general visual and language understanding tasks such as masked language modeling and masked region modeling, which enable the models to build an alignment between visual and language features. However, only feature alignments are inadequate to enhance the model's ability in conducting complex multimodal commonsense reasoning, which requires the model to understand the underlying relations and effects between objects. 

Commonsense reasoning was traditionally studied on natural language~\cite{commonsense0, commonsense1}, while recent works have paid attention to commonsense reasoning with joint visual and language inputs. 
For instance, \citet{vcr} proposes the task of Visual Commonsense Reasoning (VCR). However, the task focuses on understanding instead of generating as it asks the model to answer multiple-choice questions.
A newly introduced dataset, Visual Commonsense Generation (VCG)~\cite{vcg}, provides a more challenging task by requiring the model to generate commonsense inferences about what might happen \textit{before}/\textit{after}, and the present \textit{intents} of characters (see Table~\ref{tab:cases} for an example). In this work, we propose to tackle the task of VCG by leveraging our \textbf{K}nowledge Enhanced \textbf{M}ultimodal \textbf{BART}~\cite{bart}, which we call \textbf{KM-BART}. KM-BART is a Transformer-based model consisting of an encoder and a decoder and is pretrained on carefully designed tasks for VCG. Figure~\ref{fig:model} presents our model architecture\footnote{\url{https://github.com/FomalhautB/KM-BART-ACL}}.

Our contributions in this work are three-folded:

\begin{itemize}
    \item[1.] We extend the BART model to process multimodal data of images and texts, and enable multimodal reasoning by introducing task-relevant tokens.
    
    \item[2.] To improve the model performance on Visual Commonsense Generation (VCG), we implicitly incorporate commonsense knowledge from external knowledge graphs to our KM-BART by designing a novel pretraining task, which we call Knowledge-based Commonsense Generation (KCG). 

    \item[3.] Besides KCG, we further equip our KM-BART with standard pretraining tasks including Masked Language Modeling (MLM), Masked Region Modeling (MRM), as well as Attribution Prediction (AP) and Relation Prediction (RP). Experimental results show that all pretraining tasks are effective, and combining these pretraining tasks enable our KM-BART to achieve state-of-the-art performance on the VCG task.

\end{itemize}

\section{Related Work}

\subsection{Vision-Language Models}\label{sec:vl}
Visual-Language (VL) tasks such as Visual Question Answering (VQA)~\cite{vqa} and Image-Text Matching~\cite{it} require the models to process multimodal inputs and comprehend visual and textual information simultaneously. Inspired by successful pretrained language models like BERT~\cite{bert} and GPT-2~\cite{gpt-2}, numerous multimodal image-text pretraining and representation learning models~\cite{lxmert, vilbert, uniter, ernie-vil} have been proposed. These multimodal pretrained models use Transformers as backbone and are denoising autoencoders trained to predict the alignment of image-text pairs and the semantics of masked words and image regions.

The models mentioned above typically focus more on understanding tasks.
To further bridge the gap between visual and textual clues in multimodal data, in addition to cross-modal understanding, a model should also acquire abilities to complete generation tasks, for example, the image-to-text task of Image Captioning~\cite{caption}. However, directly transferring a model pretrained on VL understanding tasks to generation tasks is infeasible, as these models are merely Transformer-based encoders and are thus not suitable for generation tasks. 

~\citet{vlp} ease this problem by using a Transformer-based network as both an encoder and a decoder, making the model capable of generating texts based on visual and textual inputs. While~\citet{oscar} propose OSCAR, which improves the generation ability by introducing object tags as an additional clue during pretraining.  These models achieve state-of-the-art performance in downstream multimodal generation tasks such as Image Captioning~\cite{caption}.

\subsection{Commonsense Knowledge}\label{sec:common}

\begin{figure*}[!t]
    \centering
    \includegraphics[width=0.95\linewidth]{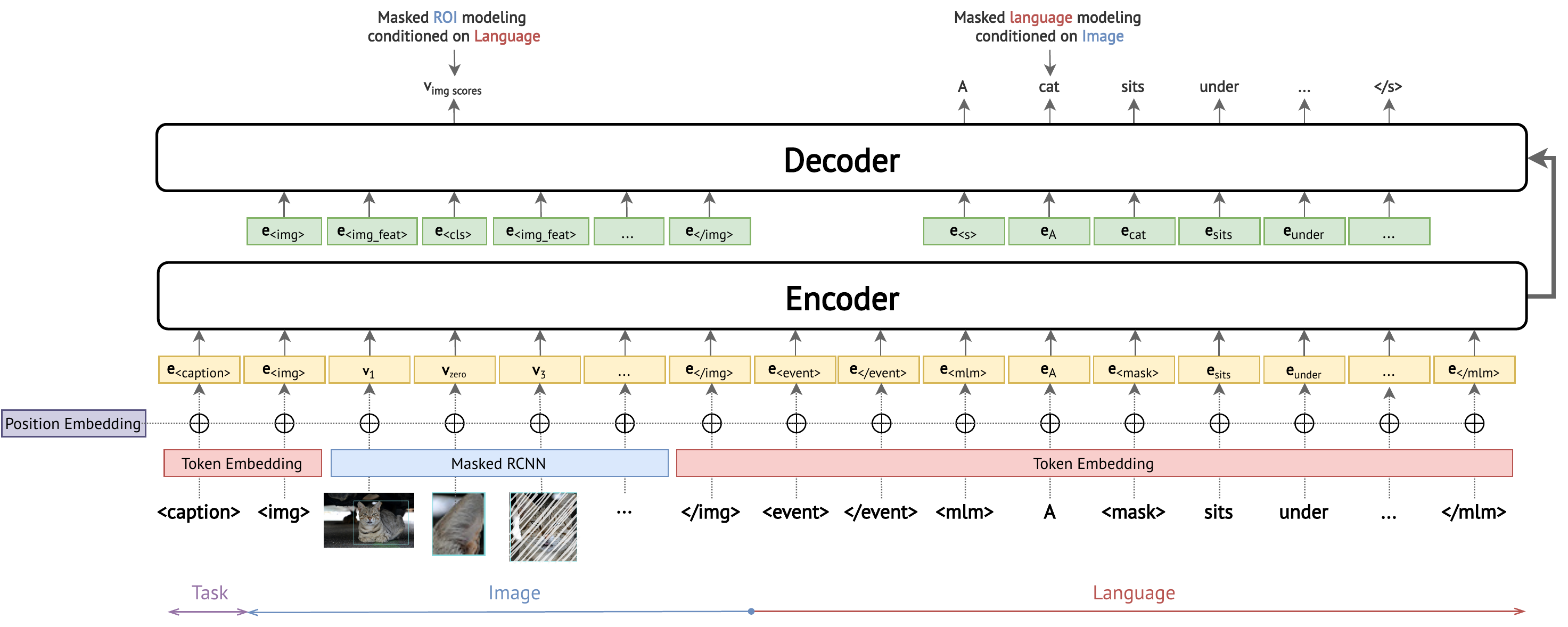}

    \caption{Model architecture. Our model is based on BART. Conditioned on prompts that indicate the task type, such as \texttt{<caption>} in the figure, our model can generate texts based on visual and textual inputs from the encoder. Our model uses different special tokens to indicate task types and inform the model of different modalities of input.}
    \label{fig:model}
\end{figure*}

Commonsense knowledge refers to the necessary level of practical knowledge and reasoning about everyday situations and events common among most people~\cite{commonsense_tutorial}. For example, one should know that ``water is for drinking" and ``sunshine makes people warm". Simple as it looks, enabling artificial intelligence to conduct commonsense reasoning has been difficult for learning-based models~\cite{commonsense_literature0}. Researchers have resorted to knowledge graphs due to their exact graph-structured representation of knowledge to overcome this problem. For example, ConceptNet~\cite{concept} is a knowledge graph with nodes representing general concepts and edges indicating relational knowledge between concepts. Another commonsense knowledge graph, ATOMIC~\cite{atomic}, extends nodes to natural language phrases, and edges to relations such as \textit{intent}, \textit{attribution}, \textit{effect}, etc. 

Despite improvements in modeling commonsense knowledge, graph-based methods require heavy human engineering, making it challenging to scale robustly. For instance, model performance usually deteriorates dramatically when retrieved contextual knowledge is noisy due to imperfect knowledge matching~\cite{kagnet}. Therefore, we implicitly leverage external knowledge using supervision signals inferred by COMET~\cite{comet}, which is a Transformer-based, generative model pretrained on commonsense knowledge graphs including ConceptNet and Atomic. Given a natural language phrase and a relation type, COMET generates natural language commonsense descriptions.

In summary, on the one hand, existing cross-modal architectures not focusing on commonsense interpretation as their pretraining tasks are designed for multimodal understanding, making them unsuitable for the downstream VCG task. On the other hand, Transformer-based generative models such as COMET~\cite{comet} cannot generate commonsense inferences from cross-modal inputs. Therefore, in this work, we propose KM-BART to conduct the task of Visual Commonsense Generation (VCG). Our KM-BART is pretrained on a dedicated pretraining task for VCG as well as other standard pretraining tasks. Experimental results show that our KM-BART achieves state-of-the-art performance on the VCG task.

\section{Methodology}\label{sec:method}

In this section, we describe our methodology for Visual Commonsense Generation. Section~\ref{sec:model} gives our model architecture. Section~\ref{sec:tasks} introduces our pretraining tasks as well as our self-training based data filtering technique. 

\begin{table}[!t]
\centering
\resizebox{0.85\linewidth}{!}{%
\begin{tabular}{lll}
\toprule
                    & \#\textbf{images} & \#\textbf{sentences} \\ \midrule \midrule
Conceptual Captions~\cite{cc} & 2,683,686 & 2,683,686    \\ \midrule
SBU~\cite{sbu}                & 780,750  & 780,750     \\ \midrule
COCO~\cite{coco}              & 82,783    & 414,113     \\ \midrule
Visual Genome~\cite{vg}       & 86,461    & 4,322,358    \\ \midrule
Total                         & 3,633,680  & 8,200,907    \\ \bottomrule

\end{tabular}
}

\caption{Statistics of pretraining datasets.}
\label{tab:dataset}
\end{table}

\subsection{Model Architecture}\label{sec:model}

Figure~\ref{fig:model} illustrates the architecture of our KM-BART. The backbone of our model is BART~\cite{bart}, which is a Transformer-based sequence-to-sequence autoencoder. We modify the original BART to adapt the model to cross-modality inputs of images and texts. We add special tokens to adapt the model to different pretraining/evaluation tasks. In the following subsections. We give the details of our visual feature extractor, the encoder, and the decoder.

\subsubsection{Visual Feature Extractor}

Following previous work on Vision-Language models~\cite{lxmert, vilbert}, we use a convolution neural network pretrained on the COCO dataset to extract visual embeddings, which are subsequently fed to the Transformer-based cross-modal encoder. Specifically, we use the pretrained Masked R-CNN~\cite{rcnn} from detectron2\footnote{\url{https://github.com/facebookresearch/detectron2}}. For each image, the pretrained Masked R-CNN proposes the bounding boxes for detected objects. The area within a bounding box is a \textbf{R}egion \textbf{o}f \textbf{I}nterest (RoI). We leverage the intermediate representations of the RoIs in the Masked R-CNN to obtain fixed-size embeddings for RoIs $V = \{v_1, \dots, v_i, \dots, v_{N}\}$, where $i$ is the index to RoIs, and $N$ is the number of RoIs for an image. The visual embedding of the $i$-th RoI $v_i$ is $\bm v_i \in \mathbb{R}^d$, where $d$ is the embedding dimension. For each of the RoIs, the Masked R-CNN also outputs the class distribution $p(v_i)$, which is later used for Masked Region Modeling.

\subsubsection{Cross-Modal Encoder}

Following~\citet{bart}, the encoder of our model is based on a multi-layer bidirectional Transformer. We introduce special tokens to adapt it to our pretraining and downstream evaluation tasks. Specifically, each example starts with a special token indicating the task type of the current example.

For our pretraining task of Knowledge-Based Commonsense Generation (see Section~\ref{sec:kcg}), we use \texttt{<before>}, \texttt{<after>}, or \texttt{<intent>} as the starting special token. For Attribution Prediction and Relation Prediction (Section~\ref{sec:ap}), we use \texttt{<region\_caption>}. Finally, for Masked Language Modeling and Masked Region Modeling, we use \texttt{<caption>}. 

Furthermore, to inform the model of different modalities of inputs, we add three sets of different special tokens: For images, we use \texttt{<img>} and \texttt{</img>} to indicate the start and the end of visual embeddings, respectively. For texts, we introduce different special tokens to distinguish between two sets of textual inputs: \textit{events} and \textit{captions}. Events are image descriptions which the model uses for reasoning about future/past events or present intents of characters in the commonsense generation task, while captions are for Masked Language Modeling, where linguistic information plays a more important role. Hence, to inform the model of these two types of textual inputs, we use \texttt{<event>} and \texttt{</event>} for events, and \texttt{<mlm>} and \texttt{</mlm>} for captions. In the following sections, we denote textual inputs of words and specical tokens by $W=\{w_1,..,w_T\}$, where $T$ is the length of textual inputs. For a token $w$, its embedding is $\bm e \in  \mathbb{R}^d$, where $d$ is the dimension of the embeddings.

\begin{table*}[!ht]
\centering
\resizebox{0.9\linewidth}{!}{
\begin{tabular}{cccc}
\toprule
\textbf{Event and image} &
  \textbf{Task} &
  \textbf{Model} &
  \textbf{Generated Sentence} \\ \midrule\midrule
2 is holding an envelope &
   &
  without event$^\mathsection$ &
  \begin{tabular}[c]{@{}c@{}}\textbf{give 1 some bad news} \\ \textbf{reassure 1} \\ \textbf{contemplate what 1 is saying to her} \end{tabular} \\ \cline{3-4} 
\multirow{8}{*}{\includegraphics[width=10cm]{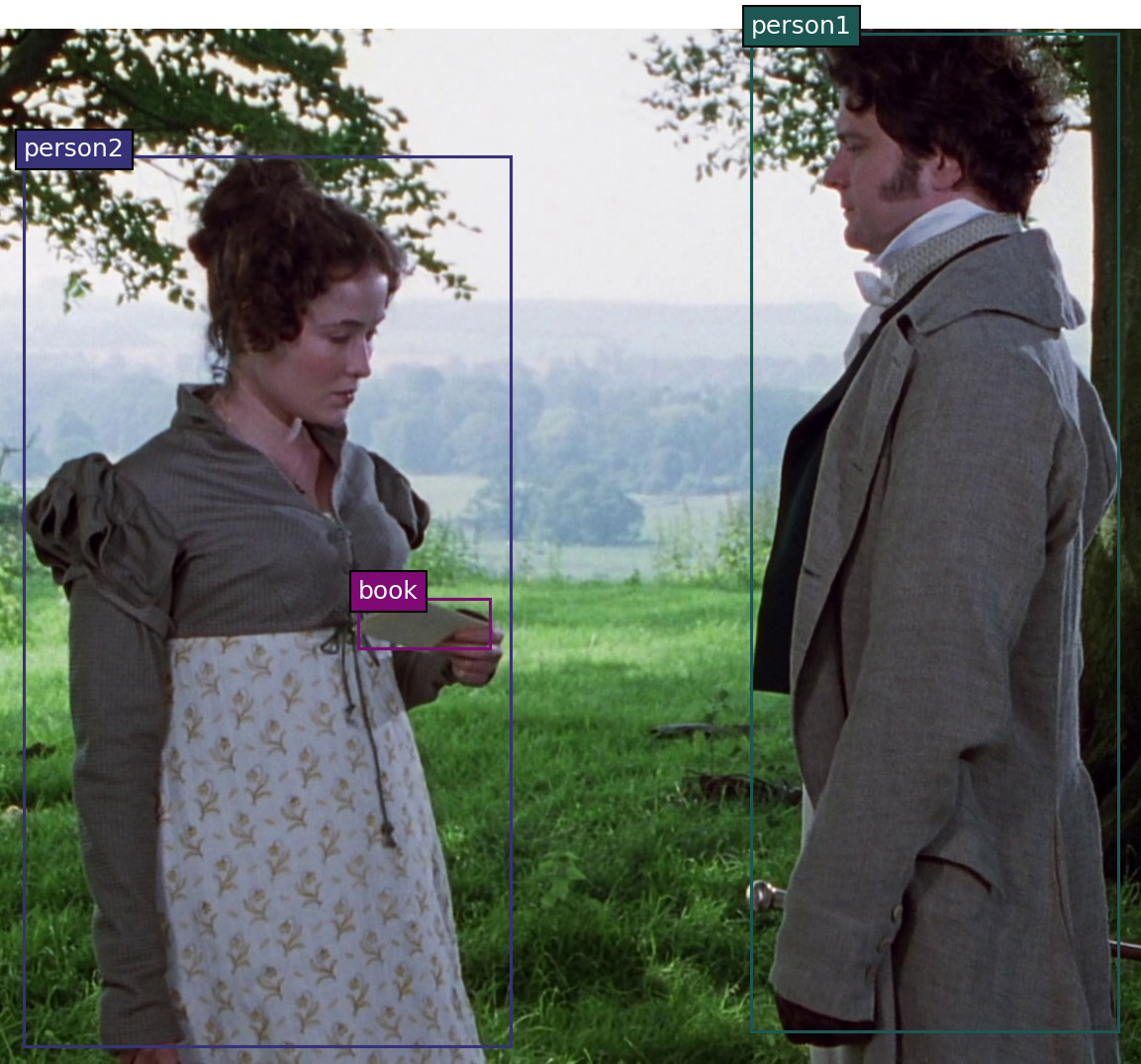}} &
  intent & 
  with event$^\dagger$ &
  \begin{tabular}[c]{@{}c@{}}\textbf{see what the letter said} \\ \textbf{give mail to 1} \\ \textbf{open the envelope} \end{tabular}  \\ \cline{3-4} 
 & 
  &
  ground truth &
  \begin{tabular}[c]{@{}c@{}}receive the envelope from 1\\ see what's inside the envelope \end{tabular} \\ \cline{2-4} 
 &
  &
  without event$^\mathsection$ &
  \begin{tabular}[c]{@{}c@{}}\textbf{walk up to 1} \\ \textbf{have seen 1 in the distance} \\ \textbf{be interested in what 1 has to say} \end{tabular} \\ \cline{3-4} 
 & 
  before &
  with event$^\dagger$ &
  \begin{tabular}[c]{@{}c@{}}\textbf{pick the envelope up} \\ \textbf{call 1 to meet him} \\ \textbf{walk to 1} \end{tabular} \\ \cline{3-4} 
 &
  &
  ground truth &
  \begin{tabular}[c]{@{}c@{}}receive mail\\ be given an envelope\\ bring the envelope with her\end{tabular} \\ \cline{2-4} 
 &
  &
  without event$^\mathsection$ &
  \begin{tabular}[c]{@{}c@{}}\textbf{finish telling 1 she has a difficult time} \\ \textbf{ask 1 what the papers are for} \\ \textbf{let go of 1} \end{tabular} \\ \cline{3-4} 
 &
  after &
  with event$^\dagger$ &
  \begin{tabular}[c]{@{}c@{}}\textbf{open the envelope} \\ \textbf{hand the envelope to 1} \\ \textbf{embrace 1} \end{tabular} \\ \cline{3-4} 
 &
  &
  ground truth &
  \begin{tabular}[c]{@{}c@{}}read the contents of the envelope to 1\\ hand the envelope to 1\\read the love letter\end{tabular} \\ \bottomrule
\end{tabular}
}
\caption{An example from the VCG dataset. We use nucleus sampling with $p=0.9$ during decoding. We show the inference sentences from (1) full model$^\mathsection$ without event descriptions but with images as inputs; (2) full model$^\dagger$ with event descriptions and images as inputs; (3) ground truth. \textbf{Bold} indicates inference sentences from our KM-BART ($^\dagger$ and $\mathsection$ indicate corresponding models in Table~\ref{tab:result}). Note that the bounding boxes are not given in the VCG dataset and are predicted by a pretrained Masked R-CNN. Additional examples are available in the Supplementary Material.}
\label{tab:cases}
\end{table*}

\subsubsection{Decoder}

The decoder of our model is also a multi-layer Transformer. Unlike the encoder, which is bidirectional, the decoder is unidirectional as it is supposed to be autoregressive when generating texts. The decoder does not take the visual embeddings as inputs. Instead, we use embeddings of the special token \texttt{<img\_feat>} to replace the actual visual embeddings. For Masked Region Modeling and Masked Language Modeling, we use \texttt{<cls>} to replace the masked regions or words (see Figure~\ref{fig:model}). The model should predict the masked words and the class distribution of the masked regions during pretraining.

\subsection{Pretraining Tasks}\label{sec:tasks}

To pretrain our model, we use four image-text datasets: Conceptual Captions Dataset~\cite{cc}, SBU Dataset~\cite{sbu}, Microsoft COCO Dataset~\cite{coco} and Visual Genome~\cite{vg}. In the remaining of this section, we use $D$ to denote the individual datasets for each of the pretraining tasks. Statistics of the datasets are given in Table~\ref{tab:dataset}. The above datasets consist of examples of parallel images and texts and are widely used in previous work~\cite{lxmert, vilbert, vlp, ernie-vil}. 

\subsubsection{Knowledge-Based Commonsense Generation}\label{sec:kcg}
The knowledge-based commonsense generation (KCG) task aims to improve the performance of KM-BART on the VCG task. We leverage knowledge induced from COMET~\cite{comet}, which is a large language model pretrained on external commonsense knowledge graphs. Given a natural language phrase and a relation as inputs, COMET generates natural language phrases as commonsense descriptions. Relations of COMET include \texttt{xIntent}, \texttt{xWant}, \texttt{xNeed}, \texttt{xReact} and \texttt{xEffect}.

We only use COMET to generate new commonsense descriptions on SBU and COCO datasets due to limits in computational power for pretraining. For each image-text pair, we use COMET to generate commonsense descriptions from the text using all five relations mentioned above. To adapt COMET generated commonsense knowledge to VCG, we consider relations \texttt{xIntent} and \texttt{xWant} from COMET as \textit{intent}, \texttt{xNeed} as \textit{before}, \texttt{xReact} and \texttt{xEffect} as \textit{after}. In this way, we generate additional commonsense knowledge for SBU and COCO datasets. The newly generated dataset has more than 3.6 million examples (Table~\ref{tab:reason_data}). However, the generated commonsense knowledge is not always reasonable as only textual information is used while the visual information is completely ignored. To ease this problem, we further filter the dataset by employing a self-training based data filtering strategy.

\noindent\textbf{Self-Training Based Data Filtering} Our strategy aims to filter the generated commonsense knowledge dataset so that the examples in the filtered dataset closely resemble the examples in the VCG dataset. To achieve this goal, we first initialize our KM-BART with BART parameters and finetune KM-BART on the VCG dataset for 30 epochs. The finetuned KM-BART already has a good performance on the VCG dataset with a CIDER score of 39.13 (see Table~\ref{tab:result}).

We then leverage this finetuned model to evaluate the quality of commonsense descriptions generated by COMET. We feed the corresponding images, texts, and relations as inputs to the finetuned KM-BART and then compute the cross-entropy (CE) loss of COMET generated commonsense descriptions. 
We observe that commonsense descriptions with a lower CE loss make more sense than those with a higher CE loss.  Notice that when computing the CE loss of the COMET generated commonsense descriptions, our KM-BART leverages both the textual inputs and the visual inputs. We provide examples of our data filtering strategy in Supplementary Material.

\begin{figure}
    \center
    \includegraphics[width=0.75\linewidth]{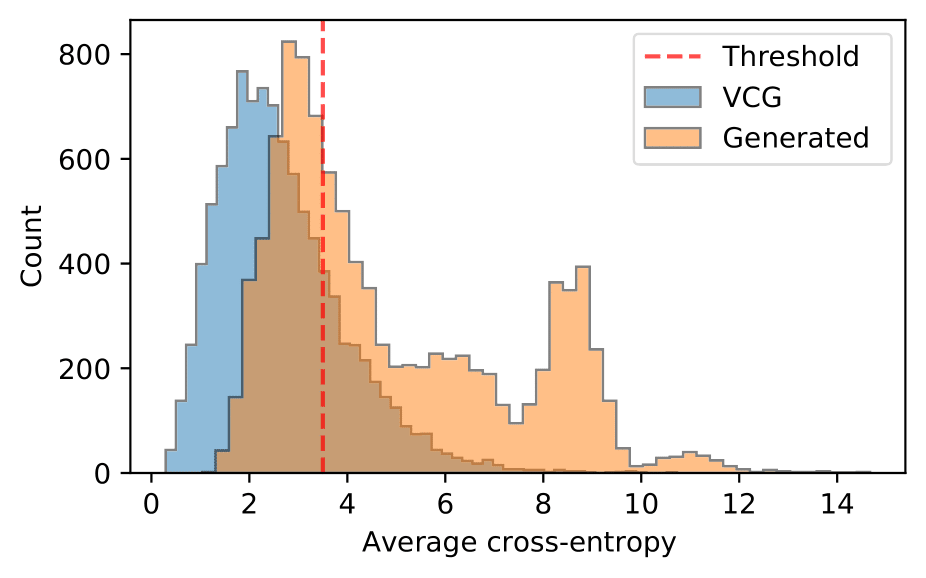}
    \caption{The distribution of the average cross-entropy on 10000 samples in the VCG dataset and our enhanced dataset. For the generated dataset, we can keep the examples of which cross entropy loss is below 3.5.}
    \label{fig:perplexity}
\end{figure}

We compute CE loss for all the commonsense descriptions in the VCG dataset and the new dataset generated by COMET. Figure~\ref{fig:perplexity} shows the distributions of CE loss for the two datasets. We observe that commonsense descriptions generated by COMET result in higher CE losses, which are expected as images are completely ignored when using COMET to generate natural language commonsense descriptions. We only keep the examples of which CE loss is below 3.5. Table~\ref{tab:reason_data} shows the statistics of generated datasets before and after data filtering. By filtering, we keep only 1.46 million examples, roughly accounting for 40\% of the original examples.

\begin{table}[!t]
\centering
\resizebox{0.7\linewidth}{!}{%
\begin{tabular}{lll}
\toprule
        & \#\textbf{Original}    & \#\textbf{Cleaned}    \\ \midrule \midrule
SBU~\cite{sbu}     & 2,032,385       &  808,425       \\ \midrule
COCO~\cite{coco}    & 1,653,075       &  660,020       \\ \midrule
Total               & 3,685,460      &   1,468,445    \\ \bottomrule
\end{tabular}
}
\caption{Statistics of datasets before and after filtering.}
\label{tab:reason_data}
\end{table}

Finally, we leverage the newly generated commonsense knowledge dataset by pretraining KM-BART on it. We expect by pretraining, the model reaches higher performance on the VCG dataset. Let ${S}=\{{w}_1,...,{w}_{L}\}$ be a commonsense description of the newly generated dataset \textit{D}, the loss function for KCG is:
\begin{equation}
\begin{split}
    & \mathcal{L}_{KCG}(\theta)=\\
    & -\mathbb{E}_{({W},{V})\sim D}\sum_{l=1}^{L}\log(P_\theta({w}_l|{w}_{<l},{W},{V}))
\end{split}
\end{equation}

\noindent where $L$ is the length of the generated sequence, $l$ is the index to individual tokens in the target commonsense description $S$, $V$ and $W$ are visual inputs and textual inputs, respectively. $\theta$ represents model parameters to be optimized.

\subsubsection{Attribute Prediction and Relation Prediction}\label{sec:ap}
The Visual Genome dataset consists of 2.3 million relationships and 2.8 million attributes. To utilize these data, we use the attribute prediction (AP) and the relation prediction (RP) as pretraining tasks, which enable the model to learn intrinsic properties among different objects in an image.

In the AP task, we feed the output vectors of the decoder for each image feature into an MLP classifier. In the RP task, we concatenate two output vectors of the decoder for each image feature pair and feed it into another MLP classifier. We use the cross-entropy loss for both tasks.

We denote the indices for AP by $1 \leq j \leq A$, the indices for RP by $1 \leq k \leq R$, where $A$ is the number of AP examples, and $R$ is the number of RP examples. We denote the label for the $j$-th AP example by $L_a(v_j)$, and the label for the $k$-th RP example as $L_r(v_{k_1}, v_{k_2})$, where $v_{k_1}$ and $v_{k_1}$ are the two RoIs of the current RP example. The loss function for the AP task is:

\begin{equation}
\begin{split}
    & \mathcal{L}_{AP}(\theta)=\\
    & -\mathbb{E}_{(W, V)\sim D}\sum_{j=1}^{A} \log(P_\theta(L_{a}(v_j) \mid W, V))
\end{split}
\end{equation}

\noindent And the loss function for the RP task is:

\begin{equation}
\begin{split}
    & \mathcal{L}_{RP}(\theta)=\\
    & -\mathbb{E}_{(W, V)\sim D}\sum_{k=1}^{R} \log(P_\theta(L_r(v_{k_1}, v_{k_2})) \mid W, V))
\end{split}
\end{equation}

\vspace{-0.2cm}

\subsubsection{Masked Language Modeling}

Following previous works~\cite{bert, roberta}, we randomly mask the input textual tokens with a probability of 15\% in the Masked Language Modeling (MLM) task. Within this 15\% of the tokens, we use \texttt{<mask>} to replace the masked token with a probability of 80\%, use a random token to replace with a probability of 10\%, and keep the masked token unchanged with a probability of 10\%.

We denote the mask indices by $1 \leq {m} \leq M$, where $M$ is the number of masked tokens. We denote the masked token by $w_m$, and the remaining tokens that are not masked by ${w}_{\setminus {m}}$, the loss function for MLM is defined as:

\begin{equation}
\begin{split}
    & \mathcal{L}_{MLM}(\theta) = \\
    & -\mathbb{E}_{(W, V)\sim D} \sum_{m=1}^{M} \log(P_\theta({w}_{m}|{w}_{\setminus {m}}, W, V))
\end{split}
\end{equation}

\subsubsection{Masked Region Modeling}

In the Masked Region Modeling (MRM) task, we sample image regions and mask the corresponding feature vectors with a probability of 15\%. The masked vector will be replaced by a vector filled with zeros. The model needs to predict the distribution over semantic classes for the masked regions. The loss function is to minimize the KL divergence of the output distribution and the distribution predicted by the Masked R-CNN used in visual features extraction. 

We denote the mask indices by $1 \leq n \leq N$, where $N$ is the number of masked regions. We let $p({v}_n)$ denote the class distribution of the masked region ${v}_n$ detected by Masked R-CNN, $q_\theta(v_n)$ denote the class distribution output by our model, the loss function for MRM is then:  

\begin{equation}
\begin{split}
    & \mathcal{L}_{MRM}(\theta) = \\
    & \mathbb{E}_{(W,V)\sim D} \sum_{n=1}^{N} D_{KL}(p({v}_n)||q_\theta({v}_n)))
\end{split}
\end{equation}

\subsubsection{Combining Losses}

To combine all the losses we described above, we weight each of the losses by $W_{KCG}, W_{AP}, W_{RP}, W_{MLM}, W_{MRM} \in \mathbb{R}$. The weights are chosen to roughly balance every term during the training phase. The final loss is:
\begin{equation}
\begin{split}
    \mathcal{L} = \ & W_{KCG} \mathcal{L}_{KCG} +  W_{AP}\mathcal{L}_{AP} + W_{RP}\mathcal{L}_{RP} + \\ 
    & W_{MLM}\mathcal{L}_{MLM} + W_{MRM}\mathcal{L}_{MRM}
\end{split}
\end{equation}

\begin{table}[!t]
\centering
\resizebox{1.0\linewidth}{!}{%

\begin{tabular}{lcccccc}
\toprule
  \textbf{Pretraining Task(s)}            & \textbf{Event} & \textbf{BLEU-2} & \textbf{METEOR} & \textbf{CIDER} & \textbf{Unique} & \textbf{Novel}\\ \midrule \midrule
\textit{\textbf{Random init}}   &            &        &        &    &        &   \\ \midrule
w/o pretraining & Y        & 22.28  & 14.55  & 36.49 & 27.81 & 29.71 \\
KCG          & Y          & 22.16  & 14.52  & 37.06 & {33.01}  & 31.20\\
KCG (before filtering)& Y & 22.24  & 14.43  & 37.08 & \textit{33.64}  & 31.37\\
AP \& RP    & Y          & 22.49  & 14.64  & 37.18 & 28.97 & 30.28 \\
MLM \& MRM  & Y          & 22.44  & 14.70  & 37.44 & 31.16 & \textit{31.64} \\ 
Full Model           &  Y        & \textbf{-} & \textbf{-} & \textbf{-}  & \textbf{-} &\textbf{-}\\
\midrule
\textit{\textbf{BART init}}     &            &        &        &       &        &\\ \midrule
w/o pretraining & Y        & 22.86  & \textbf{15.17}  & 39.13 & 27.41  & 28.32\\
{KCG}  & Y    & \textbf{23.47}  & \textit{15.02}  & \textbf{39.76}& 27.28 & 27.97 \\
KCG (before filtering)& Y & 22.90  & 14.98  & 39.01 & 26.59   & 27.13\\
AP \& RP    & Y          & 22.93  & 14.99  & 39.18 &  28.06 & 28.88\\
MLM \& MRM  & Y          & 23.13  & 14.93  & 38.75 &  28.68 & 28.74\\ 
\textbf{Full Model}$^\dagger$           &  Y        & \textit{23.25} & 15.01 & \textit{39.20}  & \textbf{35.71} & \textbf{32.85}\\

\midrule\midrule
\textit{\textbf{Random init}}   &            &        &        &      &        & \\ \midrule
w/o pretraining & N        & 13.54  & 10.14  & 14.87 & 12.19 & 24.22 \\
KCG          & N          & 13.64  & 10.12  & 15.34 & 15.95   & 25.79 \\
KCG (before filtering)& N & 13.67  & 10.13  & 15.22& 16.47 & 24.97 \\
AP \& RP    & N          & 13.83  & 10.28  & 15.48 & 14.60  & 24.75 \\
MLM \& MRM  & N          & \textit{14.36}  & \textit{10.73}  & \textit{16.72} & {15.86} & {26.12} \\ 
\textbf{Full Model}$^\mathsection$           & N         & \textbf{14.49}  & \textbf{10.86}  & \textbf{17.37} & \textit{16.89}  & {25.69}\\ \midrule
\textit{\textbf{BART init}}     &            &        &        &      &        & \\ \midrule
w/o pretraining & N        & 8.108  &  8.673 & 6.335 &  4.850 & 10.55 \\
KCG           & N         & 13.28 & 10.06 & 14.17 & 13.08 & 25.70\\
KCG (before filtering)& N& 13.29 & 10.12 & 13.93 & 13.51 & 25.59\\
AP \& RP      & N          & 12.17  & 9.503  & 12.49 & \textbf{20.98} & \textbf{29.01} \\
MLM \& MRM    & N          & 13.36  & 10.22  & 14.52 & 15.02  & \textit{28.36}\\ 
Full Model           & N         & \textbf{-}  & \textbf{-}  & \textbf{-} & \textbf{-} & \textbf{-}\\
\bottomrule
\end{tabular}
}

\caption{Results of different pretraining tasks on VCG validation set. To speed up comparison between different pretraining tasks, we use greedy decoding to generate one inference sentence per example. \textbf{Bold}: best performance. \textit{Italic}: second best performance. \textbf{Event}: whether or not event descriptions are used during \textbf{training and evaluation}.}
\label{tab:result}
\end{table}

\section{Experiments}

We describe our experiments in this section. Section~\ref{sec:settings} is the experimental settings of different pretraining and initialization strategies. Section~\ref{sec:metric} gives the evaluation task and metrics. We show our results in Section~\ref{sec:result}. In Section~\ref{sec:example}, we give example inferences generated by our model. We have the human evaluation results in Section~\ref{sec:human}.

\subsection{Settings}\label{sec:settings}

In our experiments, following the base model from~\citet{bart}, we fix the model architecture to a 6-layer encoder and a 6-layer decoder. To understand how each pretraining task helps model performance on the downstream task of VCG, we ablate on pretraining tasks. We use the following experimental settings: (1) Without any pretraining; (2) Only with Knowledge-based Commonsense Generation; (3) Only with Attribute Prediction and Relation Prediction; (4) Only with Masked Language Modeling and Masked Region Modeling; (4) With all the pretraining tasks combined. For only with Knowledge-based Commonsense Generation, we further compare the model performance before and after data filtering (see Section~\ref{sec:kcg}).

For each of the above settings, we initialize the model from random or from BART weights, respectively. Besides, we are most interested in the model performance under two settings (see the second column of Table~\ref{tab:result}): (1) Only using images as inputs; (2) Using both images and event descriptions as inputs. Note that when only using images as inputs for evaluation, we also do not use textual inputs during pretraining/finetuning.

\begin{table*}[!t]
\centering
\resizebox{0.75\linewidth}{!}{%

\begin{tabular}{lccccccc}
\toprule
                            & \textbf{Modalities} & \textbf{Event} & \textbf{BLEU-2} & \textbf{METEOR} & \textbf{CIDER} & \textbf{Unique} & \textbf{Novel}\\ \midrule \midrule

\multirow{2}{*}{\citet{vcg}$^{a\ast}$}                 & \multirow{2}{*}{\shortstack[c]{Image+Event+\\Place+Person}} &   \multirow{2}{*}{N}      & \multirow{2}{*}{\textbf{10.21}}  & \multirow{2}{*}{\textbf{10.66}} & \multirow{2}{*}{\textbf{11.86}} & \multirow{2}{*}{\textit{33.90}} & \multirow{2}{*}{\textit{49.84}}\\
                            &  &         &   &  &  &  & \\

\citet{vcg}$^{b\ast}$                 & Image      &   N     & 6.79  & 7.13 & 5.63 & 26.38 & 46.80\\
\textbf{Ours}$^\mathsection$    &   Image    &  N  & {\textit{9.04}}   & {\textit{8.33}}  & {\textit{9.12}}  & \textbf{50.75} & \textbf{52.92}\\ \midrule \midrule

\multirow{2}{*}{\citet{vcg}$^{c\ast}$}                 & \multirow{2}{*}{\shortstack[c]{Image+Event+\\Place+Person}} &   \multirow{2}{*}{Y}      & \multirow{2}{*}{\textit{13.50}}  & \multirow{2}{*}{\textbf{11.55}} & \multirow{2}{*}{\textit{18.27}} & \multirow{2}{*}{\textit{44.49}} & \multirow{2}{*}{\textit{49.03}}\\
                            &  &         &   &  &  &  & \\

\citet{vcg}$^{d\ast}$                 & {Image+Event} &   {Y}      & 12.52  & 10.73 & 16.49 & 42.83 & 47.40\\
\textbf{Ours}$^\dagger$              &     Image+Event     &             Y        & \textbf{14.21}  & \textit{11.19} & \textbf{21.23} & \textbf{57.64} & \textbf{58.22}\\ 
\bottomrule
\end{tabular}
}

\caption{Results on VCG validation set with nucleus sampling. Following~\citet{vcg}, we use nucleus sampling with $p = 0.9$ to generate five inference sentences for each example during evaluation. $^{\ast}$: we directly use evaluations from~\citet{vcg}. \textbf{Bold}: best performance. \textit{Italic}: second best performance. \textbf{Modalities}: information used during training. \textbf{Event}: whether or not event descriptions are used during \textbf{evaluation}.}
\label{tab:comparison}
\end{table*}

\subsection{Evaluation Task and Metrics}\label{sec:metric}
We evaluate our model on the recently proposed Visual Commonsense Generation (VCG) Dataset~\cite{vcg}. Given an image and a description of the event in the image, the task aims to predict events which might happen \textit{before/after}, and the present \textit{intents} of the characters in the given image.
The dataset consists of 1174K training examples and 146K validation examples. Some examples in the dataset share the same images or events, but with different inferences for events before/after or intents at present. Table~\ref{tab:cases} gives an example of the dataset. We report our model performance on the validation set as the test set is not available yet.

Besides event descriptions, the VCG dataset also provides Place and Person information for each image. Note that although~\citet{vcg} also leverages the Place and Person information for training and evaluation, we argue that such information is not generally available in normal settings, where only images and event descriptions are given. Hence, we do not use the Place and Person information in our KM-BART. 
As an additional reference, we nevertheless show in Table~\ref{tab:comparison} the best performed models from~\citet{vcg}, which also use Place and Person information.

We use three automatic evaluation metrics, including \textbf{BLEU-2}~\cite{bleu}, \textbf{METEOR}~\cite{meteor}, and \textbf{CIDER}~\cite{cider}. Following~\citet{vcg}, we also report \textbf{Unique} as the number of inference sentences unique in generated sentences divided by the total number of sentences, and \textbf{Novel} as the number of generated sentences not in the training data divided by the total number of sentences. 

\subsection{Results}\label{sec:result}

We first ablate on different pretraining tasks to understand the effect of each task. We then combine all the pretraining tasks together to train our full model. As a last step, we pick the best performed models to compare against previous state-of-the-art system~\cite{vcg}.

Table~\ref{tab:result} shows the effect of each pretraining task to our KM-BART on the VCG dataset. We can see that all our pretraining tasks help improve model performance. Most importantly, we observe that although filtering on the commonsense generation pretraining task reduces the dataset size by more than 60\%, pretraining with KCG still reaches comparable or better performance than pretraining with KCG (before filtering). This demonstrates that our self-training based filtering technique is helpful, as it helps the model reach similar or even better performance with less training data. The advantage is most evident when we initialize from BART parameters and use both images and event descriptions as inputs. Under this setting, pretraining with KCG outperforms pretraining with KCG (before filtering) in terms of all the evaluation metrics.

For using both images and event descriptions as inputs, the model performs better when initialized from pretrained BART parameters. As pretrained BART can better leverage the information in the event descriptions. Hence, to obtain our full KM-BART model for using images and events as inputs, we adopt the setting of initializing from BART parameters. Experimental results show that our full model$^\dagger$ reaches high performance on BLEU-2, METEOR and CIDER, and that the full model$^\dagger$ generates the most unique and novel inferences. 

For using only images as inputs, models initializing from random parameters outperforms those initialized from BART parameters. We argue that initializing from BART parameters results in optimization disadvantages where the model has to switch from pure textual inputs to pure visual inputs. This observation becomes evident as the model performs the worst when no pretraining is used, which indicates that the model has to entirely rely on finetuning on the VCG dataset to adapt to visual inputs. Therefore, for using only images as inputs, we obtain our full KM-BART model by initializing from random parameters. Our full model$^\mathsection$ reaches best performance on BLEU-2, METEOR and CIDER, and is the second best in terms of Unique.

In Table~\ref{tab:comparison}, we compare our full model to previous state-of-the-art~\cite{vcg}.\footnote{Note that model performance in Table~\ref{tab:comparison} is not directly comparable to that of Table~\ref{tab:result} as we use different decoding strategies to generate different number of inference sentences per example in these two tables.} We observe that although our full model$^\dagger$ taking as inputs images and event descriptions does not use Place and Person information, the model still outperforms previous state-of-the-art (\citet{vcg}$^c$). For using only images as inputs, our model$^\mathsection$ also performs better than previous results~(\citet{vcg}$^b$). Furthermore, our model$^\mathsection$ reaches comparable performance to~\citet{vcg}$^a$ in terms of BLEU-2, METEOR and CIDER, with much higher performance on Uniqueness and Novelty, even though our model$^\mathsection$ uses much less information during training compared to~\citet{vcg}$^a$.

\subsection{Case Study}\label{sec:example}

In Table~\ref{tab:cases}, we show example inferences and compare the results of our model predictions to the ground truths. The generated sentences from the model without event descriptions as inputs can already capture the most important information of commonsense. We also observe that adding event descriptions to the inputs helps the model generate more details. We gives more examples of our model in the Appendix.

\subsection{Human Evaluation}\label{sec:human}
\begin{table}[]
\centering

\resizebox{0.95\linewidth}{!}{%
\begin{tabular}{lccccc}
\toprule
\multicolumn{1}{c}{\textbf{Models}} & \textbf{Event} & \textbf{Before} & \textbf{After} & \textbf{Intent} & \textbf{Total} \\ \midrule \midrule
\citet{vcg}$^{c\ast}$   & N & 38.7          & 31.3          & 30.7          & 33.3          \\
\textbf{Ours}$^\mathsection$ & N & \textbf{61.3} & \textbf{68.7} & \textbf{69.3} & \textbf{66.7} \\ \midrule
\citet{vcg}$^{c\ast}$   & Y & 48.0          & 48.0          & 38.7          & 44.9          \\
\textbf{Ours}$^\dagger$ & Y & \textbf{52.0} & \textbf{52.0} & \textbf{61.3} & \textbf{55.1} \\ \bottomrule
\end{tabular}%
}
\vspace{-0.1cm}
\caption{Human Evaluation results. We compare the inference generated by our best model under the setting of \textit{with event} or \textit{without event}. $^\dagger$ and $\mathsection$ indicate corresponding models in Table~\ref{tab:result}. We use \citet{vcg}$^{c\ast}$ for both \textit{with event} and \textit{without event} as \citet{vcg} only release the weights of this model.}
\label{tab:human}
\end{table}

We conduct human evaluation to further understand how humans perceive the inferences generated by our KM-BART. We employ a comparison approach for a better assessment between our KM-BART and the model from~\citet{vcg}. To be specific, we randomly sample 30 examples from the VCG validation set. For each example, we use our KM-BART or the baseline model to generate 5 sets of inferences, each of which consist of the task type \textit{before}, \textit{after}, and \textit{intent}. 

We use two settings for our human evaluation: (1) With event: event descriptions are given as input during inference time;  (2) Without event: event descriptions are \textbf{not} given during inference time. Under each of the settings we compare our KM-BART model with the mode from~\citet{vcg}. We use the same 30 examples for each model under the two settings. For each example in a task type (\textit{before}, \textit{after}, or \textit{intent}), we generate 5 inferences for one model of each setting. 
In total, we generate 450 inferences for each model of each setting during the human evaluation. 

For the same example, we use our KM-BART and the model from~\citet{vcg} to generate an inference under one of the three task types, then the workers choose the more reasonable inference from the two generated inferences. We hire three workers from Amazon Mechanical Turk\footnote{\url{https://www.mturk.com/}} to evaluate each inference. We take the majority of the three workers as the final evaluation for an inference. Among all the inferences, we use the percentage of one model better than another model as the score of that model. For example, in Table~\ref{tab:human}, the score of our model (\textbf{Ours}$^\mathsection$) is 61.3 for the task type \textit{before} when event descriptions are missing. This indicates that our model is better than the baseline model for the task type \textit{before} in 61.3\% of the cases. We also take the average over the three task types as the final score (see \textbf{Total} in Table~\ref{tab:human}).

From Table~\ref{tab:human}, we can observe that our model outperforms~\citet{vcg} under both of the settings. To be specific, when event descriptions are not given, among all the inferences, our model is better than~\citet{vcg} in 66.7\% of the cases. Furthermore, our model has a lead of at least 22.6\% over~\citet{vcg} in each individual task. For example, our model generates better inferences in 68.7\% of the cases in task type \textit{after}, while the model from~\citet{vcg} is only better than our model in 31.3\% of the cases. We can obtain similar results when looking at the task type \textit{before} and \textit{intent}. 

When event descriptions are given, our model is still better than~\citet{vcg} in 55.1\% of all the cases. For each individual task, the advantage of our model is smaller when event descriptions are given than when event descriptions are not given, showing that our model can better capture information from the images.

\vspace{-0.1cm}

\section{Conclusion and Future Work}
\vspace{-0.1cm}

In this paper, we propose \textbf{K}nowledge Enhanced \textbf{M}ultimodal \textbf{BART} (\textbf{KM-BART}), which is a Transformer-based model capable of reasoning about and generating commonsense descriptions from cross modality inputs of images and texts. We propose the pretraining task of Knowledge-Based Commonsense Generation, which improves the reasoning ability of KM-BART by leveraging a large language model pretrained on external commonsense knowledge graphs. We use the self-training technique to filter the automatically generated commonsense descriptions. Experimental results on the VCG task show that our KM-BART pretrained on the pretraining tasks reaches state-of-the-art performance. Further human evaluation demonstrates that our KM-BART can generate commonsense inferences of high quality.

For future work, we plan to further expand our pretraining dataset for Knowledge-Based Commonsense Generation by including the Conceptual Captions Dataset~\cite{cc}. Furthermore, while we argue that Place and Person information is not generally available in practical scenarios, we still plan to add Place and Person information to our model in the future.

\newcommand\tableheight{8pt}

\bibliographystyle{acl_natbib}
\bibliography{main}

\clearpage
\appendix

\section{Implementation Details}

Our KM-BART is based on HuggingFace Transformers\footnote{\url{https://huggingface.co/transformers/}} and PyTorch\footnote{\url{https://pytorch.org/}}. For all our experiments, we use PyTorch built-in automatic mixed precision to speed up training. Our model has about 141 million parameters.

\subsection{Pretraining}

In pretraining, we use the AdamW optimizer with a learning rate of 1e-5. We use a dropout rate of 0.1 for regularization in fully connected layers. We pretrain our model for 20 epochs under each of the pretraining settings. We conduct our pretraining on 4 Titan RTX GPUs with an effective batch size of 256. Pretraining the model with all four pretraining tasks takes around one week. For our full model, we set the loss weights $W_{KCG}, W_{AP}, W_{RP}, W_{MRM}$ to 1.0 and $W_{MLM}$ to 5.0.

\subsection{Finetuning}

We use the same optimizer and learning rate during finetuning on the VCG dataset. We use a larger dropout rate of 0.3 as the VCG dataset is much smaller than the entire pretraining dataset. The model converges after 30 epochs. We use a single GPU with a batch size of 64. Finetuning the model takes around 40 hours.

\section{Additional Generated Examples}

Table~\ref{tab:sampled} and Table~\ref{tab:greedy} show additional examples from our model on the VCG validation set. All the commonsense inferences are generated by the best performed model.

\section{KCG Filtering Examples}

Table~\ref{tab:filtered} shows the average cross-entropy of our model on the generated COMET sentences. Lower cross-entropy indicates the generated inference sentences are more reasonable.

\begin{table}[h]
\centering
\resizebox{1.0\linewidth}{!}{%
\begin{tabular}{ccccc}
\hline
\textbf{Image} & \textbf{Event} & \textbf{Task} & \textbf{Label}                        & \textbf{cross-entropy} \\ \hline \hline
\multirow{4}{*}{\begin{tabular}[c]{@{}c@{}}\includegraphics[height=3cm]{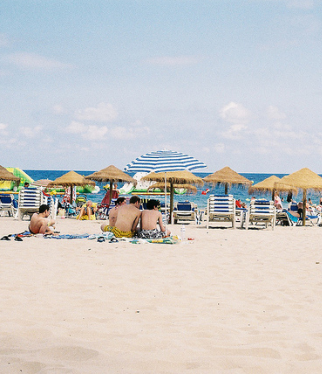}\end{tabular}} & \multirow{4}{2.5cm}{A lot of people that are at the beach}  & \addstackgap[\tableheight]{after}    & gets sunburned  & 2.755          \\ \cline{3-5}
                & & \addstackgap[\tableheight]{before}    & to drive to the beach         & 3.100          \\ \cline{3-5} 
                &  & \addstackgap[\tableheight]{intent} & to have fun     & 3.398 \\ \cline{3-5} 
                & & \addstackgap[\tableheight]{intent}    & to be safe                & 4.079         \\ \hline
\multirow{4}{*}{\includegraphics[height=3cm]{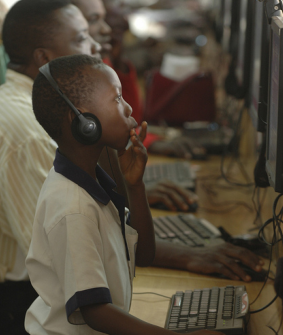}} & \multirow{4}{3cm}{Children sitting at computer stations on a long table}                                            & \addstackgap[\tableheight]{intent} & to listen to the music & 2.234 \\ \cline{3-5} 
                & & \addstackgap[\tableheight]{before}    & to have a computer       & 2.847          \\ \cline{3-5} 
                & & \addstackgap[\tableheight]{intent}    & to play with the little girl & 3.710          \\ \cline{3-5} 
                & & \addstackgap[\tableheight]{after}     & gets yelled at               & 4.055          \\ \hline
\multirow{4}{*}{\includegraphics[height=3cm]{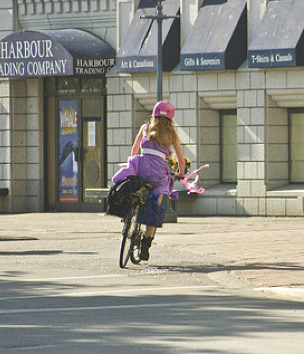}}  & \multirow{4}{3cm}{A woman is wearing a pink helmet and riding her bike through the city}                                           & \addstackgap[\tableheight]{intent} & to get to the city        & 2.255 \\ \cline{3-5} 
                & & \addstackgap[\tableheight]{after}    &   gets hit by a car       & 2.761         \\ \cline{3-5} 
                & & \addstackgap[\tableheight]{before}    & to buy a bike               & 3.052          \\ \cline{3-5} 
                & & \addstackgap[\tableheight]{after}    & gets exercise               & 4.815          \\ \hline
\multirow{4}{*}{\includegraphics[height=3cm]{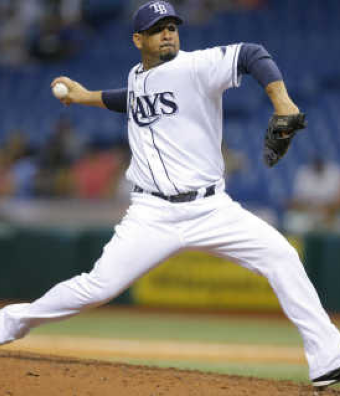}}   & \multirow{4}{3cm}{A baseball player preparing to throw a pitch during a game}                                          & \addstackgap[\tableheight]{intent} & to win the game       & 2.241 \\ \cline{3-5} 
                & & \addstackgap[\tableheight]{after}    & gets hit by a ball         & 2.773          \\ \cline{3-5} 
                & & \addstackgap[\tableheight]{before}    & to go to the stadium               & 3.222          \\ \cline{3-5} 
                & & \addstackgap[\tableheight]{intent}    & to get a tan               & 4.405          \\ \hline
\multirow{4}{*}{\includegraphics[height=3cm]{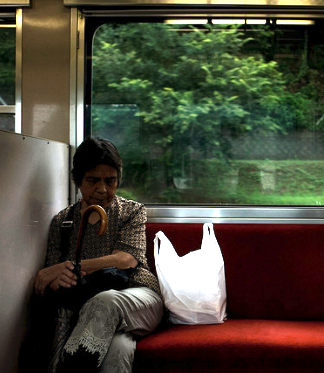}}  & \multirow{4}{3cm}{An older woman riding a train while sitting under it's window}                                           & \addstackgap[\tableheight]{before} & to go to the train station        & 1.797 \\ \cline{3-5} 
                & & \addstackgap[\tableheight]{intent}    & to get off the train        & 1.922          \\ \cline{3-5} 
                & & \addstackgap[\tableheight]{intent}    & to go to the park                & 3.232          \\ \cline{3-5} 
                & & \addstackgap[\tableheight]{after}    & refreshed               & 8.125          \\ \hline
\end{tabular}
}
\caption{Examples of commonsense descriptions generated by COMET. Examples with lower cross entropy are more reasonable. Here ``Event" refers to captions in SBU and COCO dataset. }
\label{tab:filtered}
\end{table}

\begin{table*}[ht]
\centering
\resizebox{0.9\linewidth}{!}{

\begin{tabular}{ccccc}

\multicolumn{5}{c}{Event: 1 is talking to 2 a doctor}  \\
\multicolumn{5}{c}{\includegraphics[width=10cm]{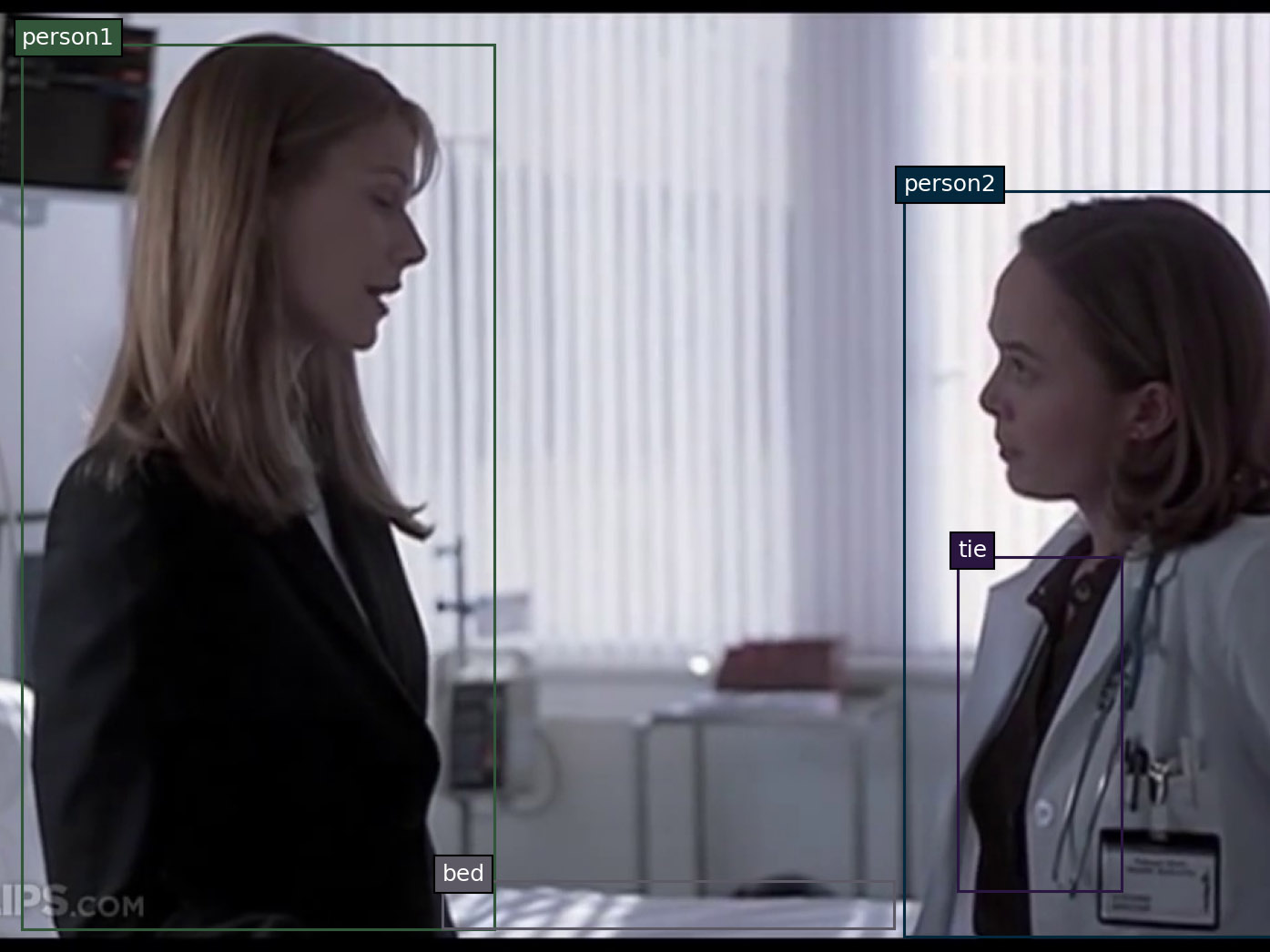}} \\
\multicolumn{5}{c}{}\\ \hline

Task & Ground Truth & Input & KM-BART & VCG \\ \hline \hline

\multirow{4}{*}{intent} & 
\multirow{4}{*}{\begin{tabular}[c]{@{}c@{}}ask 2 a question\\find out medical information \end{tabular}} & 
without event & 
\begin{tabular}[c]{@{}c@{}}\textbf{go home} \\ \textbf{say goodbye to 2} \\ \textbf{hear 2's opinion} \end{tabular} & 
\begin{tabular}[c]{@{}c@{}}make herself felt better \\ maintain the political demeanor \\ enjoy the company of his girl friend \end{tabular} \\ \cline{3-5} 

& 
& 
with event & 
\begin{tabular}[c]{@{}c@{}}\textbf{hear the doctor's diagnosis} \\ \textbf{ask the doctor some questions} \\ \textbf{get her opinion on the procedure} \end{tabular} & 
\begin{tabular}[c]{@{}c@{}}talk about her injuries \\heal his leg \\ do her job as a nurse \end{tabular} \\ \hline

\multirow{4}{*}{before} & 
\multirow{4}{*}{\begin{tabular}[c]{@{}c@{}}feel scared for her sick relative\\follow 2 into an empty room\\see 2 go into another room\end{tabular}} & 
without event & 
\begin{tabular}[c]{@{}c@{}}\textbf{take her test results} \\ \textbf{walk up to 2} \\ \textbf{enter the patient's room} \end{tabular} & 
\begin{tabular}[c]{@{}c@{}} decide on an outfit for the event \\ lose a bet \\ check his schedule to see what time it is \end{tabular} \\ \cline{3-5} 

& 
& 
with event & 
\begin{tabular}[c]{@{}c@{}}\textbf{meet 2 in the hospital} \\ \textbf{walk into the room} \\ \textbf{read a diagnosis} \end{tabular} & 
\begin{tabular}[c]{@{}c@{}}call 2 into his office \\ hear of a prescription taking \\ visit 2 in the hospital \end{tabular} \\ \hline

\multirow{4}{*}{after} & 
\multirow{4}{*}{\begin{tabular}[c]{@{}c@{}}ask 2 how bad her condition is\\tell 2 her loved one needs help\\leave the hospital\end{tabular}} & 
without event & 
\begin{tabular}[c]{@{}c@{}}\textbf{leave the hospital} \\ \textbf{walk out the door} \\ \textbf{introduce themselves to 2} \end{tabular} & 
\begin{tabular}[c]{@{}c@{}} talk about something serious with 1 \\ greet the man \\ walk away" \end{tabular}\\ \cline{3-5} 

& 
& 
with event & 
\begin{tabular}[c]{@{}c@{}}\textbf{tell 2 her symptoms} \\ \textbf{get some medicine for 2} \\ \textbf{ask 2 some questions} \end{tabular} & 
\begin{tabular}[c]{@{}c@{}}wait patiently \\ hug 2 \\ listen to the response from 2 \end{tabular} \\ \hline

\multicolumn{5}{c}{} \\
\multicolumn{5}{c}{Event: 1 is sitting at the table with 2 smoking a cigarette}  \\
\multicolumn{5}{c}{\includegraphics[width=10cm]{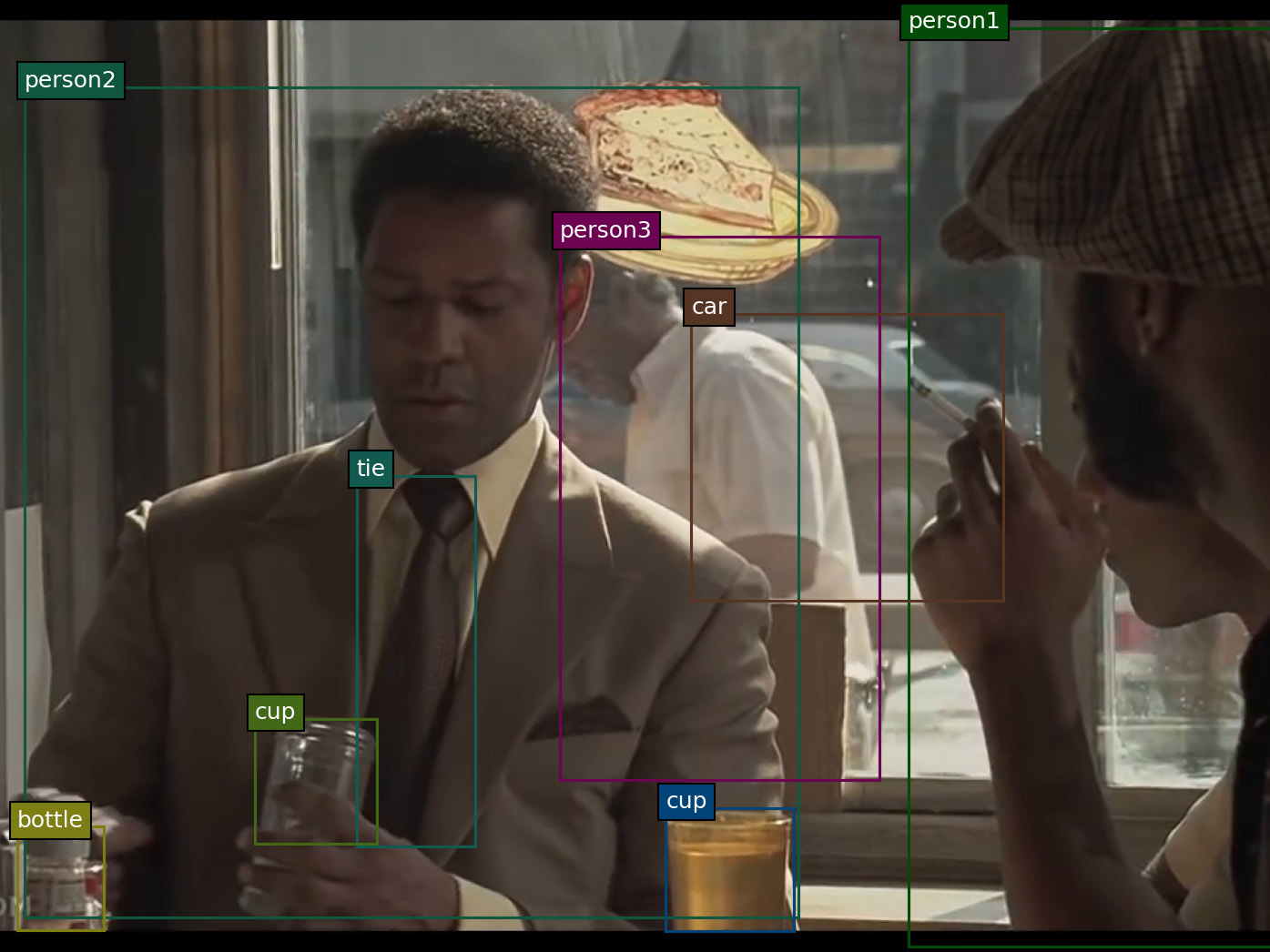}} \\
\multicolumn{5}{c}{} \\ \hline

Task & Ground Truth & Input & KM-BART & VCG \\ \hline \hline

\multirow{4}{*}{intent} & 
\multirow{4}{*}{\begin{tabular}[c]{@{}c@{}}smoke a cigarette\\talk with 2 about something \end{tabular}} & 
without event & 
\begin{tabular}[c]{@{}c@{}}\textbf{spend quality time with 2} \\ \textbf{stay at ease} \\ \textbf{speak to 2} \end{tabular} & 
\begin{tabular}[c]{@{}c@{}}have 1 shake hands \\ nod in agreement \\ do what 1 says \end{tabular} \\ \cline{3-5} 

& 
& 
with event & 
\begin{tabular}[c]{@{}c@{}}\textbf{have a smoke} \\ \textbf{get to know 2 better} \\ \textbf{get a nicotine fix} \end{tabular} & 
\begin{tabular}[c]{@{}c@{}}have lunch with 2 \\ satisfy his craving for nicotine \\ light up \end{tabular} \\ \hline

\multirow{4}{*}{before} & 
\multirow{4}{*}{\begin{tabular}[c]{@{}c@{}}order food from the waiter\\take a drink from their water cup\\be seated at a table at the restaurant\end{tabular}} & 
without event & 
\begin{tabular}[c]{@{}c@{}}\textbf{order the drink} \\ \textbf{notice 2 sitting alone at the table} \\ \textbf{enter a restaurant} \end{tabular} & 
\begin{tabular}[c]{@{}c@{}}say bye to 1 \\ sip the drink \\ look up from the food\end{tabular} \\ \cline{3-5} 

& 
& 
with event & 
\begin{tabular}[c]{@{}c@{}}\textbf{take out a cigarette} \\ \textbf{want a light} \\ \textbf{have a seat at the table} \end{tabular} & 
\begin{tabular}[c]{@{}c@{}}have 2 meet him for dinner \\ get a cigarette from 2 \\ light the cigarette\end{tabular} \\ \hline

\multirow{4}{*}{after} & 
\multirow{4}{*}{\begin{tabular}[c]{@{}c@{}}offer to help 2 get sugar for his coffee\\discuss business with 2\\watch 2 leave the restaurant\end{tabular}} & 
without event & 
\begin{tabular}[c]{@{}c@{}}\textbf{finish their meal} \\ \textbf{tell 2 something important} \\ \textbf{order lunch2} \end{tabular} & 
\begin{tabular}[c]{@{}c@{}} chat while she waits for her food \\ hug his friend \\ watch his partner 's reaction \end{tabular} \\ \cline{3-5} 

& 
& 
with event & 
\begin{tabular}[c]{@{}c@{}}\textbf{finish his meal} \\ \textbf{continue his conversation with 2} \\ \textbf{blow out smoke} \end{tabular} & 
\begin{tabular}[c]{@{}c@{}}finish smoking \\ reminisce \\ hand the cigarette to 2 \end{tabular} \\ \hline

\end{tabular}
}

\caption{Additional examples from the VCG validation set. Generated with nucleus sampling (top $p = 0.9$) .The bold texts are generated by KM-BART. We chose the KM-BART models which have the best performance, with or without event descriptions, respectively.}
\label{tab:sampled}
\end{table*}

\begin{table*}[ht]
\centering
\resizebox{0.9\linewidth}{!}{

\begin{tabular}{ccccc}

\multicolumn{5}{c}{Event: 7 is a bartender serving a customer a drink}  \\
\multicolumn{5}{c}{\includegraphics[width=10cm]{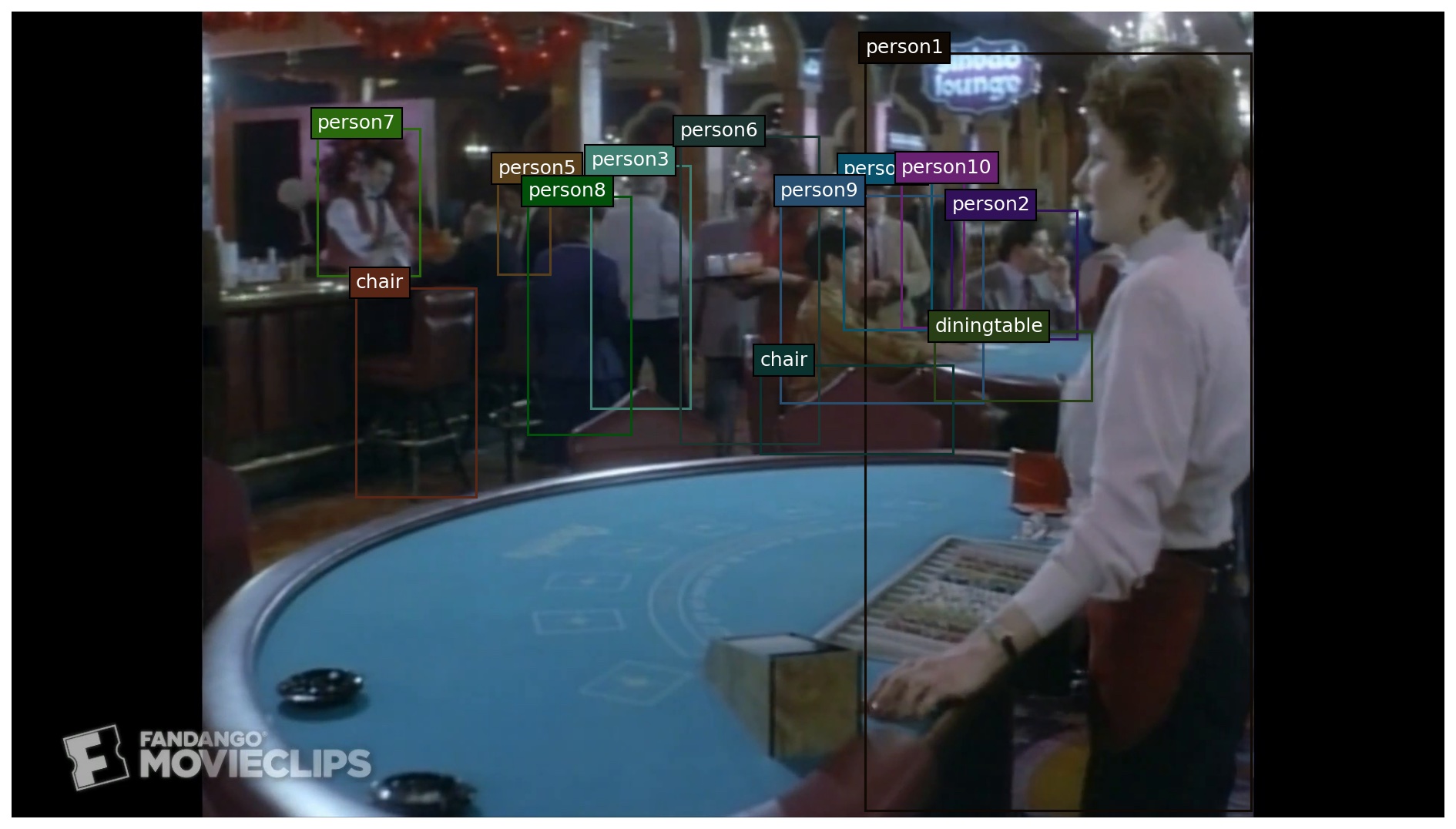}} \\ 
\multicolumn{5}{c}{}\\ \hline

Task & Ground Truth & Input & KM-BART & VCG \\ \hline \hline

\multirow{2}{*}{intent} & 
\multirow{2}{*}{\begin{tabular}[c]{@{}c@{}}make the customer happy\\ enjoy serving others \end{tabular}} & 
without event & 
\textbf{make sure the customers were happy} & 
look nice for the photo\\ \cline{3-5} 

& 
& 
with event & 
\textbf{get a good tip} & 
earn a good tip \\ \hline

\multirow{2}{*}{before} & 
\multirow{2}{*}{\begin{tabular}[c]{@{}c@{}}take a customers order \\walk out from behind the bar\end{tabular}}  & 
without event & 
\textbf{get a job as a bartender}& 
be dressed in a suit \\ \cline{3-5} 

& 
& 
with event & 
\textbf{get behind the bar} & 
take the customer 's money \\ \hline

\multirow{2}{*}{after} & 
\multirow{2}{*}{\begin{tabular}[c]{@{}c@{}}bring in the drink\\ask the customer for payment\end{tabular}} & 
without event & 
\textbf{take the drink back to the kitchen} & 
walk away from the table \\ \cline{3-5} 

& 
& 
with event & 
\textbf{ask the customer if they want another drink} & 
take money from the customer  \\ \hline

\multicolumn{5}{c}{} \\
    \multicolumn{5}{c}{Event: 2 stand in the front of the plane and faces the passengers}  \\
\multicolumn{5}{c}{\includegraphics[width=10cm]{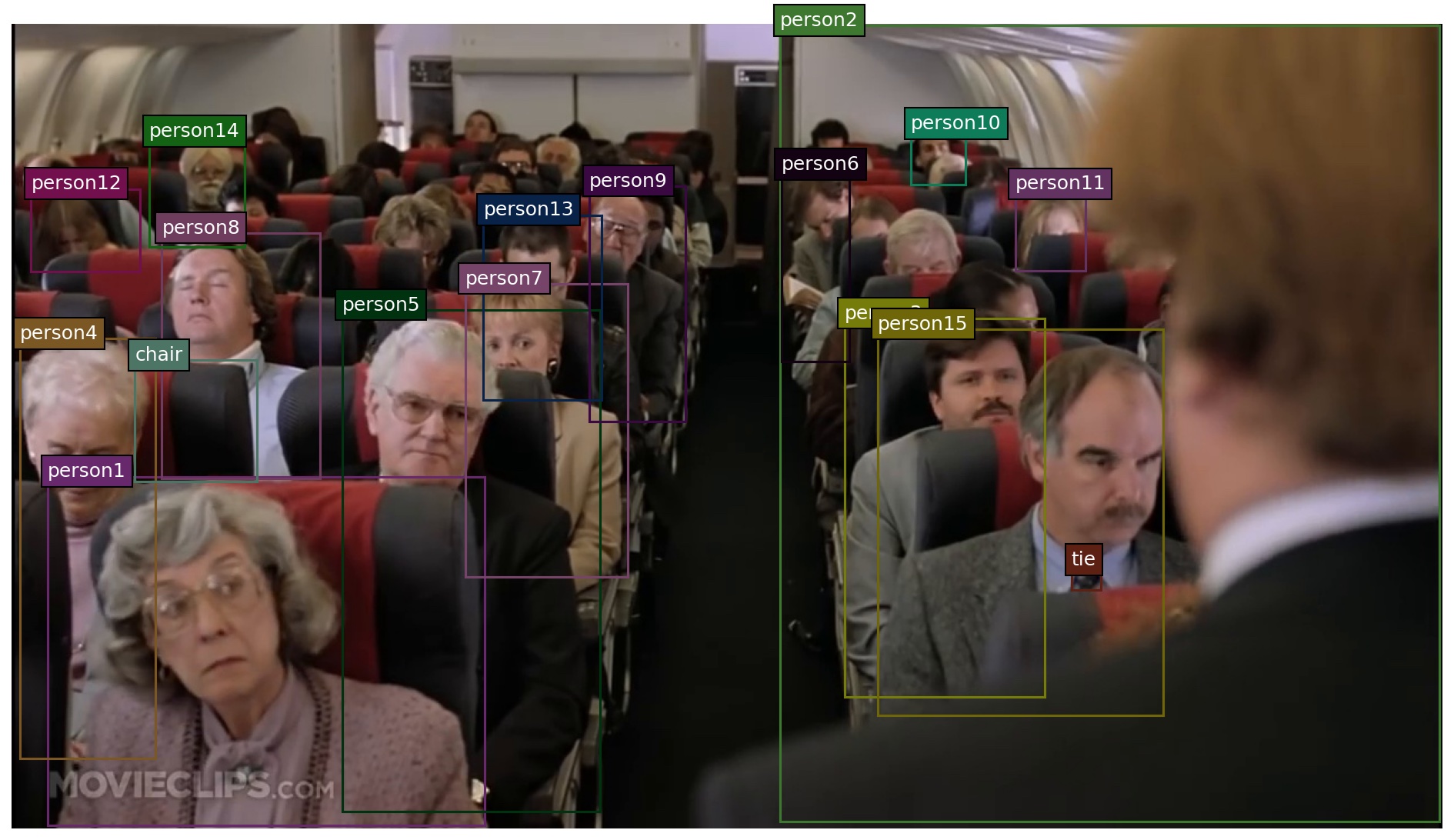}} \\ 
\multicolumn{5}{c}{} \\ \hline

Task & Ground Truth & Input & KM-BART & VCG \\ \hline \hline

\multirow{2}{*}{intent} & 
\multirow{2}{*}{\begin{tabular}[c]{@{}c@{}}make an announcement\\ tell the passengers about emergency exits \end{tabular}} & 
without event & 
\textbf{ask 1 a question} & 
see what was happening\\ \cline{3-5} 

& 
& 
with event & 
\textbf{give the passengers instructions} & 
make sure everyone had a ticket \\ \hline

\multirow{2}{*}{before} & 
\multirow{2}{*}{\begin{tabular}[c]{@{}c@{}}wait for the passengers to all take their seats \\walk to the front of the cabin \end{tabular}}  & 
without event & 
\textbf{board the plane} & 
walk into the room \\ \cline{3-5} 

& 
& 
with event & 
\textbf{walk up to the front of the plane} & 
get on the plane \\ \hline

\multirow{2}{*}{after} & 
\multirow{2}{*}{\begin{tabular}[c]{@{}c@{}}demonstrate how the exits work\\ask the passengers if they have questions\end{tabular}} & 
without event & 
\textbf{ask 1 to sit down} & 
walk away from the table \\ \cline{3-5} 

& 
& 
with event & 
\textbf{give a speech} & 
give the passengers a tour \\ \hline

\multicolumn{5}{c}{} \\
\multicolumn{5}{c}{Event: 1 holds the gun to his side looking up at the entrance to the building}  \\
\multicolumn{5}{c}{\includegraphics[width=10cm]{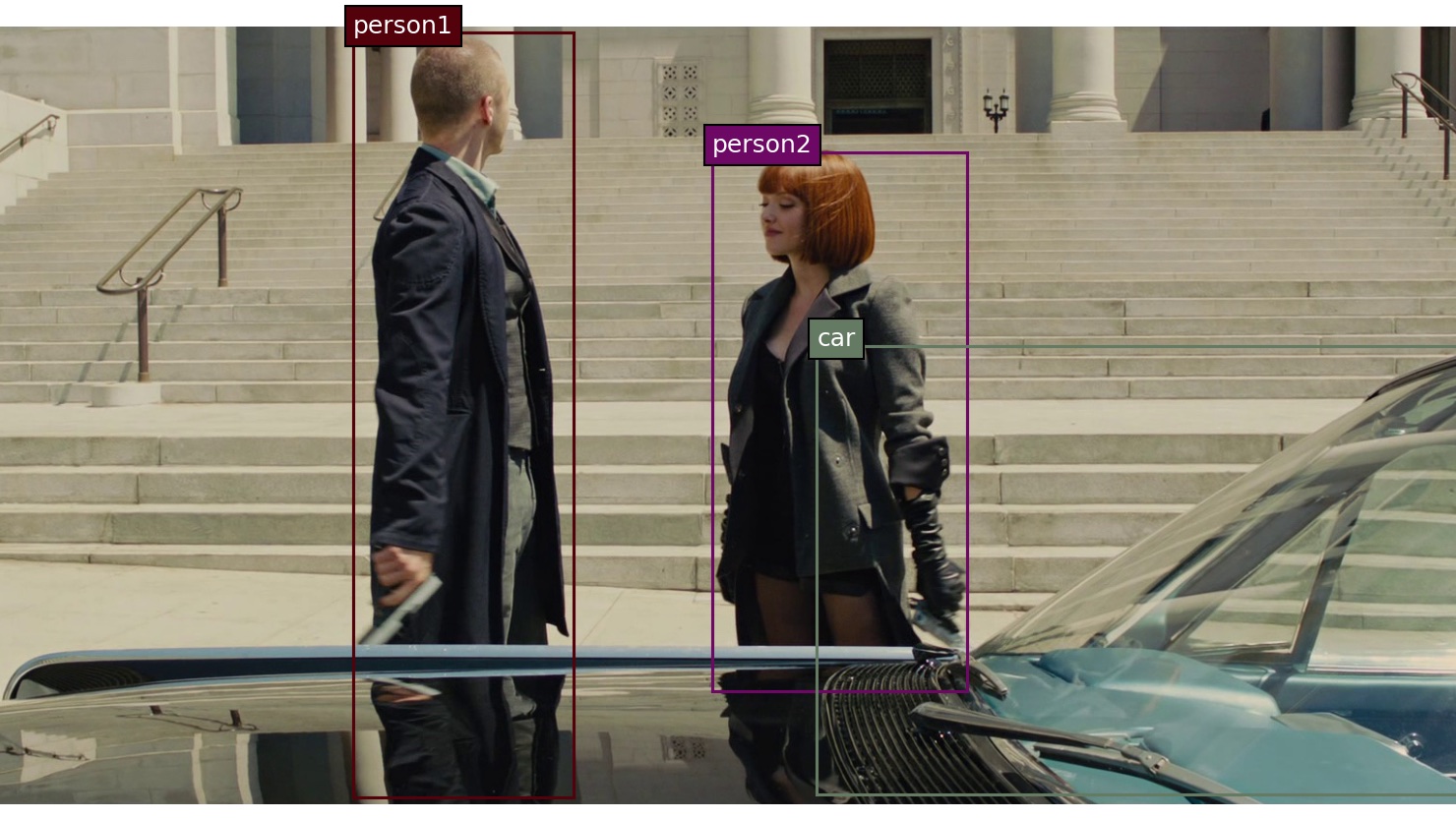}} \\ 
\multicolumn{5}{c}{} \\ \hline

Task & Ground Truth & Input & KM-BART & VCG \\ \hline \hline

\multirow{2}{*}{intent} & 
\multirow{2}{*}{\begin{tabular}[c]{@{}c@{}}scan the area for a hostile presence\\be armed for a confrontation exits\end{tabular}} & 
without event & 
\textbf{get in the car} & 
get to the car\\ \cline{3-5} 

& 
& 
with event & 
\textbf{be ready to shoot} & 
make sure no one got hurt \\ \hline

\multirow{2}{*}{before} & 
\multirow{2}{*}{\begin{tabular}[c]{@{}c@{}}draw his weapon \\drive to building to do crime \end{tabular}} & 
without event & 
\textbf{walk up to 2}& 
walk up to the car \\ \cline{3-5} 

& 
& 
with event & 
\textbf{pull out his gun} & 
get out of the car \\ \hline

\multirow{2}{*}{after} & 
\multirow{2}{*}{\begin{tabular}[c]{@{}c@{}}search for the person he wants to shoot\\enter building with gun\end{tabular}} & 
without event & 
\textbf{walk away from 2} & 
walk away \\ \cline{3-5} 

& 
& 
with event & 
\textbf{walk up to the building} & 
shoot at the entrance  \\ \hline

\end{tabular}
}

\caption{Additional examples from the VCG validation set. Generated with greedy search. The bold text are generated by KM-BART. We chose the KM-BART models which have the best performance, with or without event descriptions, respectively.}
\label{tab:greedy}
\end{table*}

\section{Additional Information on Human Evaluation}
\begin{figure*}[!t]
    \centering
    \includegraphics[width=1.0\linewidth]{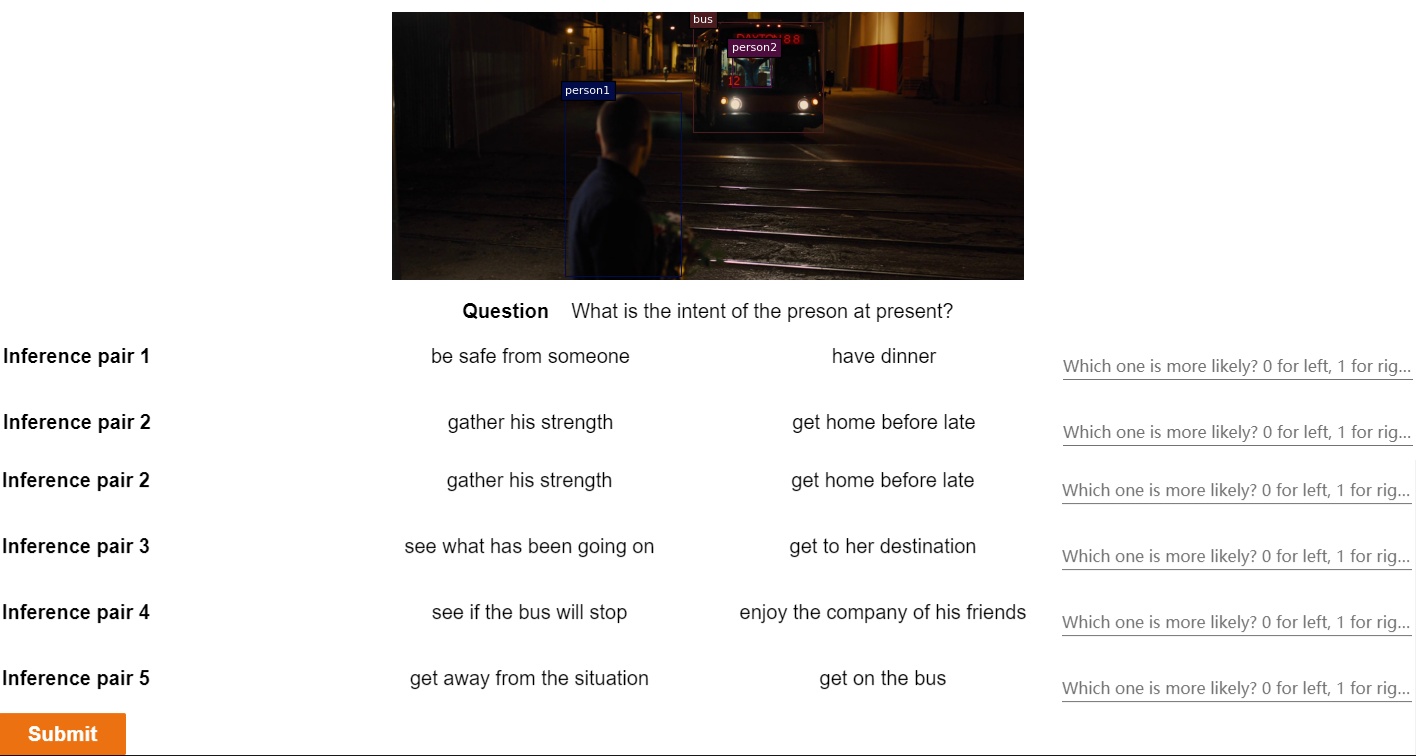}
    \caption{User interface for human evaluation.}
    \label{fig:ui}
\end{figure*}

Figure~\ref{fig:ui} is the user interface of our human evaluation. We hire workers from Amazon Mechanical Turk. We reject examples with a submission time of less than 30 seconds. The median submission time is 182 seconds. We pay for each example 0.2 USD, which is around 10.4 USD per hour.

\end{document}